\documentclass{article}
\usepackage{ijcai23}

\usepackage{times}
\usepackage{soul}
\usepackage{url}
\usepackage[hidelinks]{hyperref}
\usepackage[utf8]{inputenc}
\usepackage[small]{caption}
\usepackage{graphicx}
\usepackage{booktabs}

\usepackage{amsmath}
\usepackage{amsthm}
\usepackage{algorithm}
\usepackage{algorithmic}
\usepackage[switch]{lineno}

\usepackage{xspace}
\usepackage{textcomp} 
\usepackage{float}
\usepackage{listings}
\usepackage{caption}
\usepackage{upgreek}
\newcommand{\microf}{$\mathrm{\upmu}$-$\mathrm{F_1}$\xspace}

\newcommand{\macrof}{$\mathrm{m}$-$\mathrm{F_1}$\xspace}

\title{Breaking the Bank with ChatGPT: Few-Shot Text Classification for Finance}

\author{
Lefteris Loukas$^{1,2}$
\and
Ilias Stogiannidis$^{1,2}$\and
Prodromos Malakasiotis$^{2}$\And
Stavros Vassos$^1$
\affiliations
$^1$Helvia.ai\\
$^2$Department of Informatics, Athens University of Economics and Business, Greece\\
\emails
\{lefteris.loukas, ilias.stogiannidis, stavros\}@helvia.ai
}

\begin{document}

\maketitle

\begin{abstract}
We propose the use of conversational GPT models for easy and quick few-shot text classification in the financial domain using the Banking77 dataset. Our approach involves in-context learning with  GPT-3.5 and GPT-4, which minimizes the technical expertise required and eliminates the need for expensive GPU computing while yielding quick and accurate results. Additionally, we fine-tune other pre-trained, masked language models with SetFit, a recent contrastive learning technique, to achieve state-of-the-art results both in full-data and few-shot settings. Our findings show that querying GPT-3.5 and GPT-4 can outperform fine-tuned, non-generative models even with fewer examples. However, subscription fees associated with these solutions may be considered costly for small organizations. 
Lastly, we find that generative models perform better on the given task when shown representative samples selected by a human expert rather than when shown random ones. We conclude that a) our proposed methods offer a practical solution for few-shot tasks in datasets with limited label availability, and b) our state-of-the-art results can inspire future work in the area. 
\end{abstract}

\section{Introduction}
Virtual agents have become increasingly popular in recent years, with conversational models like GPT-3.5 \cite{instruct-gpt} and its successor ChatGPT\footnote{\url{https://chat.openai.com/}} garnering attention worldwide. While the intent detection task, as seen in the customer assistance domain, has been a well-known problem in academia for many years, it is under-explored in the financial industry due to the limited availability of datasets \cite{galitsky-ilvovsky-2019-chatbot,banking77-casanueva-etal-2020-efficient}. This study aims to bridge the gap between the financial industry and the latest developments in academia.

In this paper, we use Banking77 \cite{banking77-casanueva-etal-2020-efficient}, a real-life dataset of customer service intents and their classification labels. Unlike many datasets in the intent detection literature, Banking77 covers the niche of a single domain, contains a large number of labels (77), and many of the classes have tight overlaps between them, making it perfect for a business use-case scenario. 
Previous works have focused on fixing labeling errors \cite{ying-thomas-2022-label-banking77-1} or exploring other ways of pre-training intent representations \cite{li-etal-2022-learning-better}, which require a high level of technical expertise. 

\begin{table}[t]
    \centering
\scriptsize
\begin{tabular}{lc}
        \toprule
        \textbf{Financial Intent}  & \textbf{Label} \\
        \midrule

It declined my transfer. & Declined Transfer \\
How can I trade currencies with this app? & Exchange Via App \\
How do your exchange rates factor in? & Exchange Rate \\
I just topped up, and the app denied it. & Top-up Failed \\
There has been a red flag on my top up. & Top-up Failed \\
Tell me how to replace my expired card. & Card About to Expire\\
... & ... \\
My card is needed soon. & Card Delivery Estimate\\
What caused my transfer to fail? & Failed Transfer \\

\bottomrule
    \end{tabular}
    \caption{Example financial intents and their labels from the Banking77 dataset. In total, there are 77 different labels in the dataset.}\vspace{-2mm}
    \label{tab:example-of-banking77}
\end{table}

First, we demonstrate how well (and quickly) we can solve a few-shot financial text classification task using conversational GPT models. Secondly, we fine-tune other, non-generative, pre-trained models, based on MPNet \cite{mpnet-2020}, with SetFit \cite{setfit-Tunstall2022EfficientFL}, a recent contrastive learning technique developed by HuggingFace which aims to minimize the time and samples needed to fine-tune a pre-trained model for robust results.


Our contributions include demonstrating a clever use of in-context learning with GPT-3.5 and GPT-4 to solve a challenging intent classification task. This solution is a) especially handy when rapid and accurate results are needed for few-shot tasks in financial datasets with limited label availability, and b) requires no GPUs and minimizes the need for technical expertise, which is often lacking in the banking industry. We also show that in-context learning can perform better than fine-tuned masked language models (MLMs), even when presented with fewer examples. However, such solutions may be costly for small organizations due to subscription fees and often have limited token capacity, which only allows us to show the model 3 samples, for example. Lastly, we report state-of-the-art results by fine-tuning pre-trained models both when using the whole training dataset (Full-Data setting) and in a few-shot setting where only 10 training instances per class were used (10-shot setting) by employing SetFit and selecting representative samples after hiring a human expert.

\section{Related Work}

\subsection{Studies on Banking77}
Previous research papers provide important insights into improving the performance of financial intent classification models on the Banking77 dataset through the correction of label errors, the pre-training of intent representations, and the use of unattended tokens and example-driven training to improve utterance classification models. Initially, Casanueva et al. \shortcite{banking77-casanueva-etal-2020-efficient} established a baseline accuracy of 93.66\% by fine-tuning BERT \cite{devlin-etal-2019-bert} for the Full-Data setting, and an 85.19\% for the 10-shot setting by using a Universal Sentence Encoder \cite{use-1-cer-etal-2018-universal} and efficient Transformer representations \cite{convert-henderson-etal-2020-convert}.

                
                
                


Ying and Thomas \shortcite{ying-thomas-2022-label-banking77-1} focused on improving label errors in the Banking77 dataset and studied their negative impact on intent classification methods. They proposed two automated approaches, one through a confident learning framework ~\cite{northcutt2017rankpruning,northcutt2021confidentlearning} and one through a cosine similarity approach to identify potential label errors, which were used to flag utterances likely to be mislabeled. Experimenting in the Full-Data setting, their classifiers yielded an 88.2\% accuracy and 87.8\% F1-Score on the original dataset, and 92.4\% accuracy and 92.0\% F1-Score on the proposed trimmed dataset.

Li et al. \shortcite{li-etal-2022-learning-better} showed that pre-training intent representations can enhance the distinction of decision boundaries and improve performance in intent classification tasks in the financial domain. By using their suggested methods, prefix-tuning and fine-tuning just the last layer of a large language model (LLM), they achieved an 82.76\% accuracy and 87.35\% Macro-F1 Score on the Banking77 benchmark for the Full-Data text classification setting.

Lastly, Mehri and Eric \shortcite{mehri-eric-2021-example} proposed two approaches for improving text classification models in dialog systems: observers and example-driven training. Observers are tokens not attended to by the attention mechanism and serve as an alternative semantic representation to the [CLS] token. Example-driven training learns to classify sentences by comparing them to examples, using the underlying encoder as a sentence similarity model. Combining these approaches, Mehri and Eric reported accuracy scores of 85.95\% in the 10-shot setting and 93.83\% in the Full-Data setting.

\subsection{Few-Shot Text Classification}
Learning from just a few training instances is crucial when data collection is difficult. Interestingly, the predominant training paradigm of language model fine-tuning exhibits poor performance in few-shot scenarios \cite{dodge2020finetuning}, while the growing size of LMs often makes their use in this paradigm prohibitive. An alternative is to use in-context learning \cite{brown2020language-gpt3}, where a generative LLM is prompted with a context and is asked to solve NLP tasks without any fine-tuning. The context typically contains a short description of the task, a few demonstrations (the context), and the instance to be classified. The intuition behind in-context learning is that the LLM has already learned several tasks during its pre-training and the prompt tries to locate the appropriate one \cite{reynolds2021}. Selecting the appropriate prompt is not trivial, though; LLMs are unable to understand the meaning of the prompt \cite{webson-pavlick-2022-prompt}. This phenomenon was somewhat alleviated by fine-tuning LLMs to follow human instructions \cite{instruct-gpt,openai2023-technical-report-gpt4}. Nonetheless, in-context learning is still correlated with term frequencies encountered during pre-training \cite{razeghi-etal-2022-impact}, while instruct-based LLMs like GPT-3.5 and GPT-4 carry the biases of the human annotators that provided the training instructions. To further deal with the difficulties of in-context learning, prompt-tuning has emerged as a promising research direction \cite{prompt-tuning-1-lester-etal-2021-power,Zhou2021LearningTP-prompt-engineering,prompt-tuning-2-jiaetal-2022}.





\section{Task and Dataset}

\begin{table}[t]
\centering
\small
    \begin{tabular}{lcc}
    \toprule
    \textbf{Banking77 Statistics} & \textbf{Train} & \textbf{Test} \\
    \midrule
        Number of examples & 10,003 & 3,080 \\
        \midrule
        Minimum length in characters & 13 & 13 \\
        Average length in characters & 59.5 & 54.2 \\
        Maximum length in characters & 433 & 368 \\
        \midrule
        Minimum word count & 2 & 2 \\
        Average word count & 11.9 & 10.9 \\
        Maximum word count & 79 & 69 \\
        \midrule
        Number of intents & 77 & 77 \\
    \bottomrule
    \end{tabular}
\caption{Dataset statistics for the Banking77 dataset. The dataset contains 10,003 examples for training and 3,080 examples for testing, with 77 different intents. Text length statistics are also provided.}
\label{tab:banking77-stats}
\end{table}

Intent detection is a special case of text classification, and it plays a crucial role in task-oriented conversational systems in various domains, including finance. It reflects the complexity of real-world financial and commercial systems which can be attributed to the partially overlapping intent categories, the need for fine-grained decisions, and the usual lack of data in finance \cite{banking77-casanueva-etal-2020-efficient,loukas-etal-2021-edgar,loukas-etal-2022-finer,zavitsanos-etal-2021}. 

However, publicly available intent detection datasets are limited, and existing datasets oversimplify the task and do not reflect the complexity of real-world industrial systems \cite{simple-intent-datasets-1-braun-etal-2017-evaluating,simple-intents-coucke2018snips}. Following the recent trends towards building robust datasets for industry-ready systems \cite{better-intents-3-larson-etal-2019-evaluation,better-intent-2a-liu,better-intent-2b-Liu2021}, Banking77 \cite{banking77-casanueva-etal-2020-efficient} was created by PolyAI\footnote{\url{https://github.com/PolyAI-LDN/task-specific-datasets}} as part of their study on a new intent classifier using pre-trained dual sentence encoders based on fixed Universal Sentence Encoders \cite{use-1-cer-etal-2018-universal} and ConveRT \cite{convert-henderson-etal-2020-convert}. In contrast to other multi-domain and broad-intent datasets, which may not capture the full complexity of each domain, Banking77 is a single-domain dataset that contains a large number (77) of fine-grained intents related to banking. Casanueva et al. believe that the dataset's single-domain focus and the large number of intents make the intent detection task more realistic and challenging. However, some intent categories partially overlap with others, requiring fine-grained decisions that cannot rely solely on the semantics of individual words, indicating the difficulty of the task.

\begin{figure}[h]
  \centering
  \includegraphics[width=0.43\textwidth]{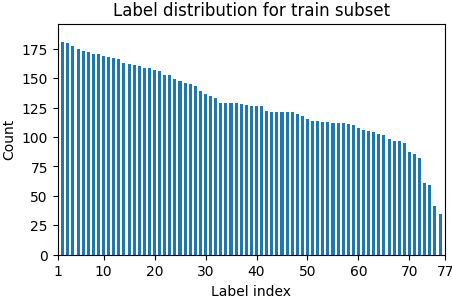}
  \caption{Class distribution of the 77 intents used over the training subset. Intent indices are shown instead of tag names for brevity.}
  \label{fig:train_subset_label_distribution}
\end{figure}

The dataset comprises 13,083 annotated customer service queries labeled with 77 intents and is split  into two subsets: train (10,003 examples) and test (3,080 samples) (Table \ref{tab:banking77-stats}). The label distribution is heavily imbalanced in the training subset (Figure \ref{fig:train_subset_label_distribution}), demonstrating the challenge in developing classifiers in the Full-Data setting.\footnote{The test subset comprises 40 instances for every label.}

\section{Methodology}


\subsection{In-Context Learning}
For in-context learning, we use \textbf{GPT-3.5} \cite{instruct-gpt} and \textbf{GPT-4} \cite{openai2023-technical-report-gpt4}, which are based on the Generative Pre-trained Transformer (GPT) \cite{radford2018improving-gpt1,radford2019language-gpt2} and further trained with Reinforcement Learning from Human Preferences (RLHF) \cite{NIPS2017_d5e2c0ad-DRLHP} to follow instructions.\footnote{\url{https://twitter.com/karpathy/status/1637147821311918083}}
GPT-3.5 is a 175B-parameter model able to consume a context o 4,096 tokens, while GPT-4 is a multi-modal model able to consume 32,768 tokens.\footnote{OpenAI has not disclosed the architecture of GPT-4.}

\subsection{Fine-tuning MLMs}

\textbf{MPNet} \cite{mpnet-2020} is a family of models based on the transformer architecture \cite{Vaswani2017AttentionIA-transformer,devlin-etal-2019-bert}, which adopts a novel pre-training objective that leverages the dependency among predicted tokens through permuted language modeling and takes auxiliary position information as input. MPNet is pre-trained on 160GB text corpora and outperforms other models like BERT \cite{devlin-etal-2019-bert}, XLNet \cite{xlnet-2019}, and RoBERTa \cite{Liu2019RoBERTaAR} on various downstream tasks. We use a variation of MPNet, establishing it as a prominent method for our task. We use two variants of MPNet, dubbed \textbf{S-MPNet-v2}\footnote{\url{https://huggingface.co/sentence-transformers/all-mpnet-base-v2}} and \textbf{P-MPNet-v2}.\footnote{\url{https://huggingface.co/sentence-transformers/paraphrase-mpnet-base-v2}}  Both variants were trained to identify similarities between pairs of texts which we believe allows the model to learn representations that encapsulate the more salient semantic details of the texts. Also, P-MPNet-v2 was trained with a more strict objective than S-MPNet-v2, which required both texts in the pair to have the exact same meaning.


\subsection{Few-Shot Contrastive Learning}
\textbf{SetFit} \cite{setfit-Tunstall2022EfficientFL} is a few-shot learning methodology that fine-tunes a pre-trained Sentence Transformer (like S-MPNet-v2) on a small number of text pairs with contrastive learning \cite{hinton-contrastivelearning-2020}. 
Tunstall et al. showed that using SetFit and eight training examples has comparable performance to training the model on the complete dataset. This could be attributed to the ability of SetFit to generate highly descriptive text representations, making it both a robust and quick solution in few-shot scenarios.

\subsection{Human Expert Annotation}
Casanueva et al. \shortcite{banking77-casanueva-etal-2020-efficient} identified class overlaps during the creation of Banking77.
To address these challenges, we curated a subset of Banking77 for few-shot text classification with the help of a human expert who reviewed a sample of 10 examples per class and selected the top 3 examples based on their relevance to the intent they represent. This approach provided a light curation that helped avoid overlaps and ensured that each example was highly relevant to its intended intent. We expect these training instances to lead to better performance than randomly selecting training instances per class in the few-shot setting.

\section{Experimental Setup}

\noindent\textbf{Fine-tuning:} For all of our methods, we use TensorFlow \cite{tensorflow2015-whitepaper} and HuggingFace \cite{wolf-etal-2020-transformers}. For the N-shot experiments using pre-trained language models and the SetFit technique, we followed the developers' recommended practices and hyperparameters.\footnote{\url{https://github.com/huggingface/setfit}}

\vspace{2mm}

\noindent\textbf{Prompt Engineering:} We experiment with two different prompt settings using GPT-4 in a 3-shot setting on a held-out validation subset.\footnote{We used 5\% of the training data.} In the first setting, we present the few-shot examples as the previous chat history. In the second setting, the few-shot examples are presented as a message from the \texttt{system}, which is one of the roles in the conversational setting of OpenAI.
The second setting yielded the best results (Table~\ref{tab:prompt_engineering}), and we proceed to use it for the rest of our experiments.


\begin{table}[h]
\small
\centering
\begin{tabular}{@{}l|cc@{}}

\toprule
Few shot examples given as & \microf & \macrof \\ \midrule
Previous chat history & 75.5 &  74.4\\
System context & \textbf{77.7} & \textbf{77.0}  \\
\bottomrule
\end{tabular} \caption{Validation Micro-F1 and Macro-F1 scores for our two prompt settings with GPT-4 in the 3-Shot scenario.} 
\label{tab:prompt_engineering}
\end{table}

\noindent\textbf{In-context Learning:} We use the OpenAI API
for the experiments with GPT-3.5 and GPT-4.\footnote{For GPT3.5, we use the \texttt{gpt-3.5-turbo} variant.}
Due to maximum token limitations, we use the 1-Shot setting for GPT-3.5 and the 3-shot setting for GPT-4.

 We instruct the model to return only one token (the financial intent label). The prompt we use can be broken down into three parts. The first contains the description of the task and the available classes, the second provides a few examples, and the third presents the text to be classified:

\begin{quote}
\scriptsize
You are an expert assistant in the field of customer service. 
Your task is to help workers in the customer service department of a company.
Your task is to classify the customer's question in 
order to help the customer service worker to answer the question. 
In order to help the worker, you MUST respond with the number 
and the name of one of the
following classes you know.
If you cannot answer the question, respond: "-1 Unknown". 
In case you reply with something else, you will be penalized.

The classes are: \\
0 activate\_my\_card\\
1 age\_limit\\
..       ..\\
75 wrong\_amount\_of\_cash\_received\\
76 wrong\_exchange\_rate\_for\_cash\_withdrawal\\

Here are some examples of questions and their classes:\\
How do I top-up while traveling? automatic\_top\_up\\
How do I set up auto top-up? automatic\_top\_up\\
...               ... \\
It declined my transfer. declined\_transfer\\

How do I locate my card?\\
\vspace{-2mm}
\end{quote}

Note that although chat models like GPT-3.5 or GPT-4 can provide a quick solution without the need for technical expertise, they come at a cost as they are only accessed behind a paywall. Our experiments cost around 60\$ when using GPT-3.5 (\$0.002 per 1K tokens) and 1,480\$ when using GPT-4 (\$0.03 per 1K tokens for the 8K context model).\footnote{\url{https://openai.com/pricing}}


\section{Results}

To comprehensively understand the model's performance, we report micro-F1 (\microf) and macro-F1 (\macrof). 
Table \ref{tab:classification_results} shows that S-MPNet-v2 achieves competitive results across all few-shot settings using SetFit. 
When trained on only 3 samples, it achieves scores of 76.3 \microf and 75.6 \macrof. 
As we increase the number of samples, the  performance improves, reaching a 91.2 micro-F1 and 91.3 macro-F1 score with 20 samples. 
This is only 3 percentage points (pp) lower than fine-tuning the model with all the data. 
Lastly, S-MPNet-v2 outperforms the previous state-of-the-art \cite{mehri-eric-2021-example}, both in the 10-shot setting (by 2.2 pp) and in the Full-Data setting (by 0.2 pp). P-MPNet-v2 has a similar but slightly worse behavior than S-MPNet-v2. 

GPT-3.5 achieves competitive results despite the fact that it is presented with only 1 sample per class (either representative or random). It outperforms S-MPNet-v2 and P-MPNet-v2 by a large margin (over 17 pp) in the 1-shot setting, while being comparable in the 3-shot setting. 
As expected, using our human-curated representative samples leads to better in-context learning results. GPT-4 also shows potential for few-shot text classification, outperforming all other models on the 3-shot setting by more than 6 pp. Similarly to GPT-3.5, its performance drops substantially (approximately 9 pp) when trained on random samples as opposed to when trained on the human-curated representative ones.



\begin{table}[t]
\centering
\small
    \begin{tabular}{lccc}
    \toprule
        Methods & Setting & \microf & \macrof \\
            \midrule
                Mehri and Eric \shortcite{mehri-eric-2021-example} & Full-Data & 93.8 & NA \\
                Mehri and Eric \shortcite{mehri-eric-2021-example} & 10-shot & 85.8 & NA \\
                Ying and Thomas \shortcite{ying-thomas-2022-label-banking77-1} & Full-Data & NA & 92.0 \\
            \midrule
                S-MPNet-v2 (ours) & Full-Data & \textbf{94.0} & \textbf{93.9} \\
                P-MPNet-v2 (ours) & Full-Data & 93.0 & 93.0 \\
            \midrule
                S-MPNet-v2 & 1-shot & 57.4 & 55.9 \\
                P-MPNet-v2 & 1-shot & 50.6 & 48.7 \\

                GPT-3.5 (representative samples) & 1-shot & \textbf{75.2} & \textbf{74.3} \\
                GPT-3.5 (random samples) & 1-shot & 74.0 & 72.3 \\
            \midrule

                S-MPNet-v2 & 3-shot & 76.3 & 75.6 \\
                P-MPNet-v2 & 3-shot & 71.4 & 70.9 \\
                GPT-4 (representative samples) & 3-shot & \textbf{83.1} & \textbf{82.7} \\
                GPT-4 (random samples) & 3-shot & 74.2 & 73.7 \\
                \midrule
                S-MPNet-v2 & 5-shot & 83.5 & 83.3 \\
                S-MPNet-v2 & 10-shot & 88.0 & 87.9 \\
                S-MPNet-v2 & 15-shot & 90.6 & 90.5 \\
                S-MPNet-v2 & 20-shot & 91.2 & 91.3 \\
            \midrule
                P-MPNet-v2 & 5-shot & 79.2 & 79.1 \\
                P-MPNet-v2 & 10-shot & 85.7 & 85.8 \\
                P-MPNet-v2 & 15-shot & 88.4 & 88.4 \\
                P-MPNet-v2 & 20-shot & 90.1 & 90.0 \\
                
    \bottomrule
    \end{tabular}
\caption{Classification results for all models on the test data, with N-Shot indicating the number of samples used during training. The MPNet and Paraphrase-MPNet models are fine-tuned without the \texttt{SetFit} method on the Full-Data setting.}
\label{tab:classification_results}
\vspace{-2mm}
\end{table}

\section{Conclusion and Future Work}

We presented a few-shot text classification study on the financial domain. Experimenting with Banking77, a financial intent classification dataset, we showed that in-context learning with conversational LLMs can be a straightforward solution when one needs fast and accurate results in few-shot settings. In addition, we demonstrated that generative LLMs, like GPT-3.5 and GPT-4, can perform better than MLM models, even when presented with fewer examples. While such solutions minimize the technical expertise needed or omit GPU training times, they can be considered costly for small organizations, given that LLMs can be only accessed behind a paywall (approximately 1,600\$ for GPT-3.5 and GPT-4).

On the other side, by fine-tuning S-MPNet-v2 with SetFit, we surpassed the previous state-of-the-art in the 10-shot setting by 2 pp. The same model also achieved state-of-the-art results in the Full-Data setting with standard fine-tuning.

In future work, we plan to experiment with more Sentence Transformers versions and other generative models like LLaMA \cite{touvron2023llama}, Alpaca \cite{alpaca} and Claude\footnote{\url{https://www.anthropic.com/index/introducing-claude}}. Lastly, it would be interesting to investigate other, more time-expensive techniques suited for few-shot text classification like PEFT \cite{peft}.

\subsubsection{Acknowledgements}

This work has received funding from European Union's Horizon 2020 research and innovation programme under grant agreement No 101021714 ("LAW GAME").

\appendix





\bibliographystyle{named}
\bibliography{ijcai23}

\end{document}